\documentclass[11pt]{article}

\usepackage[spanish,english]{babel}
\usepackage{arxiv}
\usepackage{algpseudocode}
\usepackage{algorithm}
\usepackage{graphicx}
\usepackage[english]{babel}
\usepackage{amsmath}
\usepackage{longtable}
\usepackage{siunitx}
\usepackage{booktabs}
\usepackage{subcaption}
\usepackage{adjustbox}
\usepackage{etoolbox} 
\usepackage[utf8]{inputenc} 
\usepackage[T1]{fontenc}    
\usepackage{url}            
\usepackage{booktabs}       
\usepackage{amsfonts}       
\usepackage{nicefrac}       
\usepackage{microtype}      
\usepackage{lipsum}
\usepackage{color}
\usepackage{amssymb}
\usepackage{float}
\usepackage{multirow}
\usepackage{afterpage}

\usepackage{booktabs} 
\usepackage{multirow} 
\usepackage{makecell} 
\usepackage{geometry} 
\usepackage{lineno}
\usepackage{xspace}
\usepackage{doi}
\usepackage{datetime}

\title{ChronoVAE-HOPE: Beyond Attention - A Next-Generation VAE Foundation Model for Specialized Time Series Classification}

\author{
José Alberto Rodríguez\\
  {\small Department of Computer Science}\\
  {\small and Artificial Intelligence}\\
  {\small DiCITS, iMUDS, DaSCI}\\
  {\small University of Granada, Granada, Spain 18071}\\
  \texttt{josealberto99@correo.ugr.es} \\
\And
  Luis Balderas\\
  {\small Department of Computer Science}\\
  {\small and Artificial Intelligence}\\
  {\small DiCITS, iMUDS, DaSCI}\\
  {\small University of Granada, Granada, Spain 18071}\\
  {\small Advanced Medical Imaging Group,}\\
  {\small Instituto de Investigación Biosanitaria de Granada (ibs.Granada)}\\
  \texttt{luisbalru@ugr.es} \\
\And
  Miguel Lastra \\
  {\small Department of Software Engineering}\\
  {\small DiCITS, iMUDS, DaSCI}\\
  {\small University of Granada, Granada, Spain 18071}\\
  \texttt{mlastral@ugr.es} \\ 
\And
  Antonio Arauzo-Azofra \\
  {\small Department of Rural Engineering,}\\
  {\small DiCITS, iMUDS,}\\
  {\small University of Córdoba, Córdoba, Spain 14005}\\
  \texttt{arauzo@uco.es} \\ 
\And
  José M. Benítez\\
  {\small Department of Computer Science}\\
  {\small and Artificial Intelligence}\\
  {\small DiCITS, iMUDS, DaSCI}\\
  {\small University of Granada, Granada, Spain 18071}\\
  {\small Advanced Medical Imaging Group,}\\
  {\small Instituto de Investigación Biosanitaria de Granada (ibs.Granada)}\\
  \texttt{J.M.Benitez@decsai.ugr.es} \\
}

\begin{document}

\maketitle

\begin{abstract}
Time Series Foundation Models (TSFMs) have become a new component of the state-of-the-art in general time series forecasting. However, adapting them to specialized classification tasks remains constrained by two interconnected challenges: the quadratic cost of standard attention mechanisms and the inability to disentangle the structural components underlying time series variability. This technical report introduces ChronoVAE-HOPE, a next-generation TSFM that reconciles massive generalization with structured latent representation for time series classification. The core of the proposal is a Variational Autoencoder (VAE) framework built upon the HOPE Block, which replaces quadratic attention with a dual-memory system: Titans modules for dynamic short-term retention and a Continuum Memory System (CMS) for the abstraction of long-term historical context.

A key architectural novelty is the disentangled latent space, which factorizes representations into independent trend and seasonal components via dedicated encoder heads and separate decoder pathways. ChronoVAE-HOPE undergoes self-supervised pre-training on the Monash archive, combining a Masked Time Series Modeling (MTSM) auxiliary objective with a disentangled VAE reconstruction loss. The pre-trained encoder is subsequently frozen and used to generate fixed-length embeddings for downstream classification on the UCR benchmark datasets. Empirical results demonstrate strong performance across diverse temporal domains, particularly in settings characterized by strict causal structure. ChronoVAE-HOPE establishes a robust and interpretable framework for the adaptation of foundation models to time series classification through structured generative representations.
\end{abstract}


\section{Introduction}

The capacity to extract meaningful, transferable representations from temporal data
constitutes one of the most persistent open challenges in machine learning for time
series. Applications as diverse as monitoring physiological signals in clinical
environments \cite{khan2025tracking}, fault detection in industrial sensor networks
\cite{8926446}, and regime identification in quantitative finance
\cite{10.1145/3729531} share a common demand: models must simultaneously capture
slow-varying global structure and rapid local dynamics, often across sequences of
highly variable length and statistical nature. This dual requirement has historically
resisted a unified solution because the two timescales operate through mechanisms
that standard architectures address separately and at considerable computational cost.
 
Early work on time series classification relied on similarity-based approaches. Elastic
distance measures such as Dynamic Time Warping (DTW) \cite{berndt1994using} offered
alignment-invariant comparison but scaled poorly to large corpora, while
shapelet and dictionary methods \cite{lines2012shapelet} provided interpretable
discriminative patterns at the price of expensive extraction procedures. The
transition to deep representation learning, epitomized by Fully Convolutional Networks
(FCN) \cite{wang2017time} and the multi-scale receptive fields of InceptionTime
\cite{ismail2020inceptiontime}, delivered substantial accuracy gains by learning
invariant local features end-to-end. Yet these models operate under an inherently
closed-world assumption: each dataset requires independent optimization from
randomly initialized weights, and no structural knowledge propagates across problem
domains.
 
The emergence of large-scale pre-training paradigms in natural language processing
opened a new chapter for temporal data. Patch-based encoders such as PatchTST
\cite{nie2022time} demonstrated that segmenting a time series into fixed-length tokens
preserves local semantic coherence while reducing the attention receptive field. At
industrial scale, systems like TimesFM \cite{das2023decoder} and MOIRAI
\cite{woo2024unified} showed that massive, heterogeneous pre-training yields
generalizable forecasting priors without task-specific supervision. On the generative
side, Variational Autoencoder approaches have attracted growing interest for their
ability to impose probabilistic structure on the latent space: the explicit
factorization of the posterior encourages disentangled representations that are both
interpretable and transferable, properties that purely discriminative encoders do not
guarantee by construction.
 
Despite this progress, two fundamental bottlenecks continue to obstruct the
specialization of TSFMs for classification. The first is computational: the standard
self-attention mechanism exhibits quadratic growth with sequence length, capping the
historical context that can be economically encoded in a single forward pass. The
second is representational: dominant architectures collapse all temporal variation
into a single, entangled latent vector, conflating the smooth macro-trends and the
oscillatory seasonal patterns that jointly characterize most real-world signals. This
conflation limits the structural interpretability and cross-domain transferability of
the learned embeddings, because a downstream classifier must implicitly unravel the
component structure that the encoder was never trained to separate.
 
To address these gaps, this paper introduces ChronoVAE-HOPE, a next-generation Time
Series Foundation Model that places structured generative representation at the center
of its design. ChronoVAE-HOPE contributes three interconnected innovations. First,
the HOPE Block \cite{behrouz2025nestedlearningillusiondeep} substitutes quadratic
attention with a dual-memory architecture (Titans fast-weight modules for short-term
dynamics and the Continuum Memory System (CMS) for multi-timescale historical
abstraction) achieving linear computational cost without sacrificing depth of context.
Second, a disentangled VAE bottleneck factorizes the latent space into orthogonal
trend and seasonal subspaces via dedicated encoder heads and independent decoder
pathways, supervised by an explicit series decomposition signal. Third, a composite
self-supervised pre-training objective over the Monash repository
\cite{godahewa2021monash} (combining VAE reconstruction with a Masked Time Series
Modeling (MTSM) auxiliary loss) equips the frozen encoder with embeddings
sufficiently rich for lightweight downstream classifiers trained on the UCR benchmark
\cite{8894743}.
 
The remainder of this document details the architecture, the pre-training
strategy, and the empirical evaluation, demonstrating how ChronoVAE-HOPE
establishes a principled, generative paradigm for structured latent adaptation in
time series classification.

\section{Our Proposal}

This section introduces ChronoVAE-HOPE, a foundation model for time series classification based on a disentangled Variational Autoencoder built upon the HOPE dual-memory encoder. The architecture is designed to address the inherent challenges of TSFMs: variable sequence lengths, non-stationarity, and the need to model both local semantics and long-range historical dependencies while explicitly separating structural temporal components.

\subsection{Architecture}

ChronoVAE-HOPE processes sequences of diverse natures through a meticulously designed pipeline comprising a embedding stem, a HOPE-based encoder that projects sequences into a disentangled latent space, and a structured decoder that independently reconstructs trend and seasonal components. At inference time, the frozen encoder generates fixed-length embeddings for downstream classification.

\begin{figure}[h]
    \centering
    \includegraphics[width=1.0\textwidth]{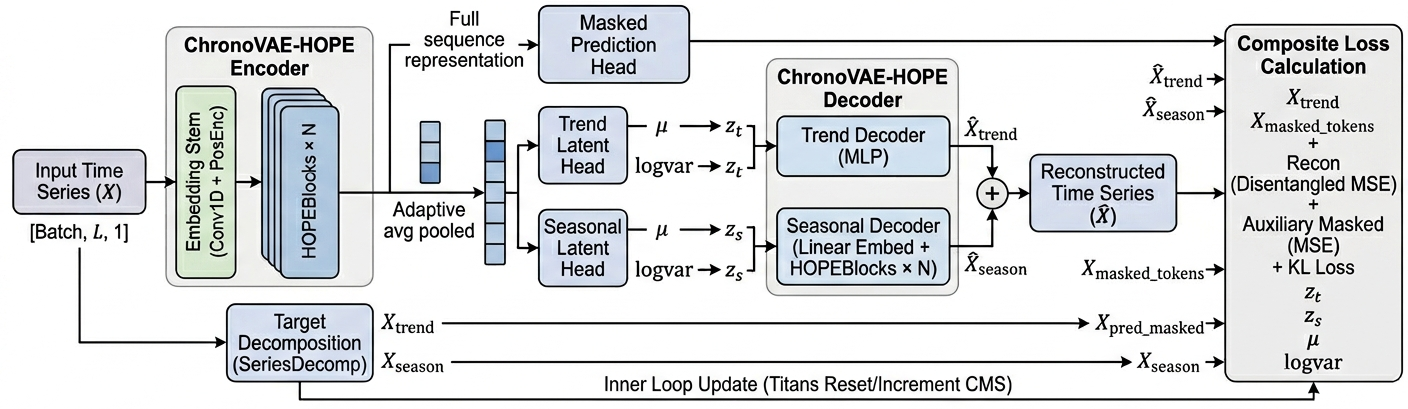}
    \caption{ChronoVAE-HOPE Backbone Architecture}
    \label{fig:chronovaehope-architecture}
\end{figure}

\subsubsection{Embedding Stem}

Before entering the encoder, the raw input sequence $X \in \mathbb{R}^{L \times 1}$ is processed by a multi-scale \textit{EmbeddingStem} module that captures local temporal patterns and introduces positional awareness. The stem applies a 1D convolution with kernel size 7 and padding 3 to project the input from its original dimensionality into the model embedding space $\mathbb{R}^{D}$, followed by a GELU activation. The resulting feature map has shape $\mathbb{R}^{L \times D}$, preserving the original sequence length. A learnable positional encoding $\mathbf{P} \in \mathbb{R}^{1 \times L_{\max} \times D}$ is then added to inject temporal position information, and a dropout layer is applied for regularization.

This convolutional stem serves a dual purpose: it captures local, multi-scale patterns through its receptive field without the need for explicit patching, while the learnable positional encoding enables the model to adapt its notion of temporal order across diverse series lengths.

\subsubsection{HOPE Encoder and latent space}

The embedded sequence is then processed by a stack of HOPE Blocks (detailed in Section~\ref{sec:hope}), producing a deep, contextualized representation $\mathbf{H} \in \mathbb{R}^{L \times D}$. To obtain a fixed-length vector suitable for the VAE bottleneck, adaptive average pooling is applied over the temporal dimension, yielding $\mathbf{h} \in \mathbb{R}^{D}$.

The central architectural novelty of ChronoVAE-HOPE is its \textbf{disentangled latent space}. Rather than projecting into a single latent distribution, the encoder maintains two independent sets of projection heads: one for the trend component and one for the seasonal component:

\begin{align}
    \boldsymbol{\mu}_t,\, \log\boldsymbol{\sigma}^2_t &= \text{FC}_{\mu,t}(\mathbf{h}),\; \text{FC}_{\sigma,t}(\mathbf{h}) \in \mathbb{R}^{Z} \\
    \boldsymbol{\mu}_s,\, \log\boldsymbol{\sigma}^2_s &= \text{FC}_{\mu,s}(\mathbf{h}),\; \text{FC}_{\sigma,s}(\mathbf{h}) \in \mathbb{R}^{Z}
\end{align}

Latent samples are drawn via the reparameterization trick:
$$\mathbf{z}_t = \boldsymbol{\mu}_t + \boldsymbol{\epsilon} \odot \exp\!\left(\tfrac{1}{2}\log\boldsymbol{\sigma}^2_t\right), \quad \mathbf{z}_s = \boldsymbol{\mu}_s + \boldsymbol{\epsilon} \odot \exp\!\left(\tfrac{1}{2}\log\boldsymbol{\sigma}^2_s\right)$$

where $\boldsymbol{\epsilon} \sim \mathcal{N}(\mathbf{0}, \mathbf{I})$. This explicit factorization encourages the model to learn orthogonal representations: $\mathbf{z}_t$ captures slow-varying macro-trends, while $\mathbf{z}_s$ encodes oscillatory or periodic dynamics. The final embedding used for downstream classification is the concatenation $\mathbf{z} = [\mathbf{z}_t \| \mathbf{z}_s] \in \mathbb{R}^{2Z}$, or equivalently $[\boldsymbol{\mu}_t \| \boldsymbol{\mu}_s]$ at inference time for a deterministic representation.

\subsection{HOPE Encoder Block}
\label{sec:hope}

Traditional Transformers \cite{wolf-etal-2020-transformers} rely on self-attention \cite{vaswani2023attentionneed}, a mechanism that incurs quadratic memory cost and struggles with long or infinite-horizon sequences. To address these limitations, ChronoVAE-HOPE replaces standard attention with the HOPE Block, a dual-memory system that integrates two advanced neural memory primitives:

\begin{itemize}
    \item \textbf{Self-Modifying Titans}: This module acts as the model's short-to-medium-term working memory. The Titans \cite{behrouz2024titanslearningmemorizetest} module maintains fast-weight matrices $\mathbf{M}_k, \mathbf{M}_v, \mathbf{M}_{\text{main}} \in \mathbb{R}^{D \times D}$ that are dynamically updated at each time step via a stabilized Delta Gradient Descent (DGD) rule on an L2 regression objective. This enables the network to adapt to sudden volatility and local temporal dynamics without maintaining an exhaustive attention matrix over the full sequence. The fast weights are updated with an exponential decay factor $\alpha = 0.99$ and a learning rate $\eta = 0.01$, and are clamped to $[-5, 5]$ for numerical stability. In parallel, slow-weight MLPs provide a meta-learned prior that complements the fast-weight retrieval.

    \item \textbf{Proper Continuum Memory System (CMS)}: Operating in conjunction with the Titans module, the CMS is responsible for long-term abstraction. It comprises $L_{\text{CMS}}$ hierarchical levels, each implemented as a two-layer MLP with GELU activation and LayerNorm. The levels operate at different update frequencies $[1, 4, 16, 64]$ batches, implementing a multi-timescale memory consolidation loop. Each level maintains a persistent memory snapshot $\mathbf{m}_i \in \mathbb{R}^{D}$, updated via exponential moving average with consolidation rates $0.1 / (i+1)$, which decrease for deeper (slower) levels. Gated mixing between fast activations and slow persistent memories at each level allows the model to selectively rely on long-range historical context when local signals are insufficient, granting ChronoVAE-HOPE a substantially extended effective receptive field.
\end{itemize}

Each HOPE Block applies the two memory modules sequentially with residual connections and LayerNorm:
$$\mathbf{x}' = \mathbf{x} + \text{Dropout}(\text{Titans}(\text{LN}(\mathbf{x})))$$
$$\mathbf{x}'' = \mathbf{x}' + \text{Dropout}(\text{CMS}(\text{LN}(\mathbf{x}')))$$

Together, the Titans and CMS modules map the input embeddings into deep, contextualized representations with linear computational complexity in the sequence length.

\subsubsection{Series Decomposition}

To provide a supervisory signal aligned with the disentangled latent space, ChronoVAE-HOPE incorporates a \textit{SeriesDecomp} module during training. This module applies a 1D average pooling filter (kernel size 25, stride 1, same padding) to extract the trend component as a moving average, and obtains the seasonal component as the residual. The decomposed components $X_{\text{trend}}, X_{\text{seasonal}}$ serve as reconstruction targets for the respective decoder branches, providing an inductive bias that aligns structural semantics in the latent space with the corresponding generative pathway.

\subsubsection{Structured Decoder}

ChronoVAE-HOPE employs two independent decoder branches that reconstruct the series from the latent vectors:

\begin{enumerate}
    \item \textbf{Trend Decoder}: A lightweight two-layer MLP that maps $\mathbf{z}_t \in \mathbb{R}^Z$ directly to the full reconstructed trend sequence $\hat{X}_{\text{trend}} \in \mathbb{R}^{L \times 1}$. This pathway is designed for the smooth, low-frequency nature of trend signals, which do not require the sequential inductive bias of the HOPE architecture.

    \item \textbf{Seasonal Decoder}: A HOPE-based pathway that first embeds $\mathbf{z}_s$ into a sequence of shape $\mathbb{R}^{L \times D}$ via a linear expansion, then applies a stack of HOPE Blocks to decode complex oscillatory patterns, and finally projects back to the original dimensionality via a linear head. This recurrent-style decoding via HOPE is particularly well-suited for the structured, periodic nature of seasonal components.
\end{enumerate}

The final reconstruction is the element-wise sum of both branches: $\hat{X} = \hat{X}_{\text{trend}} + \hat{X}_{\text{seasonal}}$.

\subsection{Training Strategy}

The training of ChronoVAE-HOPE is governed by a two-stage paradigm: a self-supervised pre-training phase on a massive heterogeneous corpus, followed by a lightweight downstream adaptation phase using the frozen encoder.

\subsubsection{Pre-training ChronoVAE-HOPE}

Pre-training does not require external labels and is driven by a composite self-supervised objective combining two complementary loss terms.

\begin{enumerate}
    \item \textbf{Disentangled VAE Reconstruction Loss.} The primary objective optimizes the Evidence Lower Bound (ELBO) separately for each structural component. The reconstruction targets are obtained from the \textit{SeriesDecomp} module applied to the original (unmasked) series. The reconstruction loss is the sum of Mean Squared Errors across both branches:
    
    $$\mathcal{L}_{\text{recon}} = \frac{1}{L}\sum_{t=1}^{L} \left\|\hat{X}^t_{\text{trend}} - X^t_{\text{trend}}\right\|^2 + \left\|\hat{X}^t_{\text{seasonal}} - X^t_{\text{seasonal}}\right\|^2$$

    The KL divergence regularization is computed independently for each component and averaged over the batch:
    
    $$\mathcal{L}_{\text{KL}} = -\frac{1}{2}\sum_{i}\left(1 + \log\sigma^2_{t,i} - \mu^2_{t,i} - \sigma^2_{t,i}\right) - \frac{1}{2}\sum_{i}\left(1 + \log\sigma^2_{s,i} - \mu^2_{s,i} - \sigma^2_{s,i}\right)$$

    \item \textbf{Masked Time Series Modeling (MTSM).} Drawing on advances in masked autoencoders, a 20\% of the input values are randomly set to zero before entering the encoder, while the full, unmasked series serves as the reconstruction target. An auxiliary linear head applied to the full encoder output sequence $\mathbf{H} \in \mathbb{R}^{L \times D}$ is trained to reconstruct the masked values:
    
    $$\mathcal{L}_{\text{MTSM}} = \frac{1}{|\mathcal{M}|} \sum_{(t, c) \in \mathcal{M}} \left\|\hat{X}_{t,c} - X_{t,c}\right\|^2$$

    where $\mathcal{M}$ is the set of masked positions. This objective encourages the encoder to develop a robust internal representation that can infer missing context from surrounding dynamics.
\end{enumerate}

The total pre-training loss is a weighted combination of all objectives:

$$\mathcal{L}_{\text{total}} = \mathcal{L}_{\text{recon}} + \lambda_{\text{KL}} \cdot \mathcal{L}_{\text{KL}} + \lambda_{\text{MTSM}} \cdot \mathcal{L}_{\text{MTSM}}$$

with $\lambda_{\text{KL}} = 0.01$ and $\lambda_{\text{MTSM}} = 0.5$.

\subsubsection{Downstream Classification}

Once the HOPE encoder has developed a structured understanding of temporal dynamics, adaptation to a target classification task is performed by freezing the encoder and training a lightweight downstream classifier. This protocol ensures that the rich, generalizable representations learned during pre-training are not degraded by task-specific gradient updates.

At inference time, the encoder operates deterministically, using the posterior means $\boldsymbol{\mu}_t$ and $\boldsymbol{\mu}_s$ instead of sampled latent vectors. The fixed-length embedding $\mathbf{z} = [\boldsymbol{\mu}_t \| \boldsymbol{\mu}_s] \in \mathbb{R}^{2Z}$ is then passed to the downstream classifier, which consists of a LayerNorm, a hidden linear layer with GELU activation, dropout, and a final linear projection to the number of target classes. Only the classifier parameters are updated during this phase, using a cross-entropy loss.

\section{Empirical Analysis}

To validate the efficacy of ChronoVAE-HOPE and its generalization capability from a self-supervised environment to specific classification tasks, a thorough experimental framework has been designed. This section details the architectural configuration, optimization hyperparameters, datasets utilized, and the evaluation scheme for the results.

\subsection{Metrics}
 
The evaluation of ChronoVAE-HOPE spans two distinct operational phases, each
requiring a different family of metrics.
 
During the \textbf{self-supervised pre-training phase}, the model is monitored
through three complementary loss-derived quantities that jointly reflect the quality
of both generative and reconstructive objectives:
 
\begin{itemize}
    \item \textbf{Disentangled Reconstruction MSE ($\mathcal{L}_{\text{recon}}$)}:
    Measures the fidelity of the structured decoder in separately reproducing the
    trend and seasonal components produced by the \textit{SeriesDecomp} module. A
    low value confirms that the latent codes $\mathbf{z}_t$ and $\mathbf{z}_s$
    retain sufficient information to reconstruct the respective temporal
    sub-structures, validating the alignment between the disentangled representation
    and its corresponding generative pathway.
 
    \item \textbf{KL Divergence ($\mathcal{L}_{\text{KL}}$)}: Quantifies the
    regularization pressure applied to both posterior distributions, measuring their
    aggregate divergence from the standard Gaussian prior. This term governs the
    tension between latent expressiveness and structural regularity, and its
    trajectory during training reflects whether the model converges to an organized,
    well-separated latent space or collapses toward a degenerate posterior.
 
    \item \textbf{MTSM Reconstruction MSE ($\mathcal{L}_{\text{MTSM}}$)}: Evaluates
    the auxiliary reconstruction accuracy of the randomly masked input positions from
    the full encoder output sequence $\mathbf{H} \in \mathbb{R}^{L \times D}$. A
    decreasing trajectory indicates that the encoder progressively develops
    contextual representations capable of inferring missing dynamics from surrounding
    temporal evidence.
\end{itemize}
 
For the \textbf{downstream classification phase}, two standard supervised metrics are
reported across each UCR benchmark dataset:
 
\begin{itemize}
    \item \textbf{Accuracy}: The proportion of correctly classified instances in the
    held-out test set, providing a primary measure of per-dataset predictive
    performance.
 
    \item \textbf{Macro-averaged F1-Score}: The unweighted mean of per-class F1
    scores, computed independently of class frequency. This metric is particularly
    informative in the presence of class imbalance, which is prevalent in several
    UCR benchmark datasets, and provides a more complete picture of the classifier's
    discriminative ability across all target categories.
\end{itemize}

\subsection{Network Hyperparameters}

The base model architecture was configured to balance expressive capacity with computational efficiency. The input sequence length is set to 256 time steps. The convolutional stem projects the input into a model embedding space of dimension $D = 128$, using a 1D convolution with kernel size 7. The encoder depth comprises 3 HOPE blocks, each operating with a Titans chunk size of 32 tokens. Regarding the memory architecture, the CMS utilizes 4 hierarchical levels with update frequencies $[1, 4, 16, 64]$ batches. A dropout probability of 0.1 is applied uniformly within the HOPE blocks and the downstream classifier head. The loss weighting coefficients are $\lambda_{\text{KL}} = 0.01$ and $\lambda_{\text{MTSM}} = 0.5$.

\subsection{Training Hyperparameters}

The optimization process utilizes Adam as the primary optimizer. During pre-training, the model is trained for up to 300 epochs with a learning rate of $1 \times 10^{-3}$ and a batch size of 256. An early stopping criterion monitors the training loss with patience 5 and minimum improvement $\delta = 0.5$.

For the downstream classification phase, the classifier is trained for up to 300 epochs using Adam with a learning rate of $1 \times 10^{-3}$ and a batch size of 256. Early stopping is applied with patience 10 and minimum improvement $\delta = 10^{-3}$.

\subsection{Datasets}

To ensure that the model assimilates a wide variety of temporal dynamics, two distinct data collections are employed for the foundation and specialization phases. The pre-training corpus consists of massive, forecasting-oriented datasets extracted from the Monash Time Series Forecasting Repository \cite{godahewa2021monash}, specifically \textit{m4\_hourly, weather, electricity, traffic}, and \textit{tourism\_monthly}. This selection exposes the model to patterns ranging from high-frequency and extreme volatility to low-frequency seasonal macro-trends. For the fine-tuning corpus, the BakeOff classification benchmark \cite{Ruiz2021}, based on 86 datasets from the UCR Time Series Classification Archive \cite{8894743}, served as the reference for downstream evaluation. To align this evaluation with typical and pragmatic scenarios in both industrial and academic contexts, the experiment focused on a filtered subset of problems satisfying two conditions: possessing fewer than eight classes and containing sequences with fewer than 400 time steps per instance.

\subsection{Results}

The following section presents the experimental results for ChronoVAE-HOPE. As shown in the appendix, Table \ref{results}, the proposed model was evaluated using the Bake Off benchmark datasets, achieving a mean test accuracy of 61.1\% and an F1-score of 0.583. When restricted to datasets containing fewer than eight classes and fewer than 400 time steps per instance, the model achieved a mean accuracy of 76.52.\% and an F1-score of 0.741. Table \ref{type}, which presents the mean accuracy categorized by dataset type using the aforementioned filter, indicates that SPECTRO datasets yielded the highest performance (85.3\%), followed by SIMULATED (82.45\%) and HAR (79.33\%) datasets. Conversely, MOTION datasets exhibited the lowest performance (66.31\%).

A more detailed breakdown of the results reveals several factors that influence 
model performance. As shown in Table \ref{class}, binary classification tasks 
yielded considerably higher results than multi-class problems, suggesting that 
the latent representations learned by the VAE are more easily separable under a 
reduced label space. Regarding time series length, Table \ref{length} shows that 
the model performed consistently well across short and medium-length series, but 
degraded markedly for longer sequences, which is consistent with the known 
difficulty of encoding long-range temporal dependencies within a fixed-dimensional 
latent space. Table \ref{ts} explores the effect of training set size, 
though no clear monotonic trend emerges: performance peaks for the smallest 
training sets and declines for moderate sizes, and the partial recovery observed 
for larger sets is difficult to attribute solely to data abundance, as training 
size is likely correlated with other dataset characteristics such as the number 
of classes or series length.

\begin{table}[]
\centering
\begin{tabular}{l S}
\hline
{\textbf{Type}} & {\textbf{Average Accuracy}}                              \\ \hline
{DEVICE}        & { 67.42}                                            \\ \hline
{ ECG}           & { 75.31} \\ \hline
{ HAR}           & { 79.33}                                            \\ \hline
{ IMAGE}         & { 74.03}                                            \\ \hline
{ MOTION}        & { 66.31}                                            \\ \hline
{ SENSOR}        & { 78.58} \\ \hline
{ SIMULATED}     & { 82.45}                                            \\ \hline
{ SPECTRO}       & { 85.30} \\ \hline
\end{tabular}
\caption{ChronoVAE-HOPE results by dataset type}
\label{type}
\end{table}

\begin{table}[]
\centering
\begin{tabular}{l S S}
\hline
{\textbf{Number of Classes (C}} & {\textbf{Mean Accuracy}}   & {\textbf{Mean F1-Score}}                           \\ \hline
{$C=2$}        & { 71.50} & {0.70}                                            \\ \hline
{ $C>2$}           & { 55.04} & {0.52} \\ \hline

\end{tabular}
\caption{ChronoVAE-HOPE results dividing datasets depending on the number of classes}
\label{class}
\end{table}

\begin{table}[]
\centering
\begin{tabular}{l S S}
\hline
{\textbf{Time Series Length (L)}} & {\textbf{Mean Accuracy}}   & {\textbf{Mean F1-Score}}                           \\ \hline
{$L \leq 80$}        & { 72.31} & {0.70}                                            \\ \hline
{ $80<L\leq136$}           & { 69.24} & {0.68} \\ \hline
{ $136<L\leq 235$}           & { 73.62} & {0.70} \\ \hline
{ $L> 235$}           & {53.89} & {0.51} \\ \hline
\end{tabular}
\caption{ChronoVAE-HOPE results dividing datasets depending on the time series length}
\label{length}
\end{table}

\begin{table}[]
\centering
\begin{tabular}{l S S}
\hline
{\textbf{Train Size (TS)}} & {\textbf{Mean Accuracy}}   & {\textbf{Mean F1-Score}}                           \\ \hline
{$TS \leq 30$}        & { 71.00} & {0.70}                                            \\ \hline
{ $30<TS\leq 70$}           & { 61.90} & {0.60} \\ \hline
{ $70<TS\leq 400$}           & { 52.17} & {0.49} \\ \hline
{ $TS> 400$}           & { 67.64} & {0.65} \\ \hline

\end{tabular}
\caption{ChronoVAE-HOPE results dividing datasets depending on the train size}
\label{ts}
\end{table}

\section{Discussion}

The analysis of the empirical results obtained from the UCR benchmark following the pre-training and frozen-encoder adaptation phases provides critical insights into the inductive biases of the model. Generally, the explicit disentanglement of trend and seasonal representations within the VAE bottleneck, combined with the HOPE dual-memory encoder, is expected to demonstrate notable advantages in domains governed by strict temporal causality, such as HAR, Sensor, and ECG datasets, where the separation of macro-trend and oscillatory dynamics is semantically meaningful.

The CMS hierarchical consolidation loop enables the model to leverage multi-timescale dependencies across its levels, which is particularly beneficial for datasets with slow-varying global structure (such as physiological signals or mechanical sensor readings) while the Titans fast-weight mechanism provides the local adaptability needed for volatile or event-driven series. This is reflected in the results: all of which exhibit structured temporal dynamics well-aligned with the model's inductive biases.
Contrary to initial expectations, IMAGE-type datasets achieved a mid-range mean accuracy of 74.03\%, surpassing MOTION (66.31\%) and DEVICE (67.42\%) categories. While the causal inductive biases of both the Titans module and the CMS are architecturally misaligned with the cyclic and direction-agnostic nature of geometric contours encoded in image-derived time series, the disentangled latent space may still capture sufficient low-frequency structural variation to support classification above chance. This suggests that the series decomposition module, despite lacking semantic interpretability in this domain, provides a degree of implicit regularization that partially compensates for the representational mismatch.

The lowest performance was observed in MOTION and DEVICE datasets, which may reflect the high intra-class variability and sensor heterogeneity characteristic of these domains, posing a more fundamental challenge to the frozen-encoder adaptation paradigm than the geometric nature of IMAGE series.

This contrast underscores that the success of the architecture depends critically on the alignment between the model's structural inductive biases and the intrinsic nature of the target domain, while also suggesting that the regularization imposed by structured disentanglement can yield partial robustness even under representational mismatch.

\section{Conclusions}
 
In this technical report, ChronoVAE-HOPE has been presented, a foundation model
designed to advance time series classification through the principled combination of
a disentangled Variational Autoencoder and the HOPE dual-memory architecture. The
empirical results validate the central hypothesis: factorizing the latent space into
orthogonal trend and seasonal subspaces, each governed by a dedicated encoder head
and an independent generative decoder, enables the learning of structured,
probabilistically grounded representations that monolithic encoders cannot produce
by construction.
 
The primary conclusions derived from this development are summarized as follows:
 
\begin{itemize}
    \item \textbf{Computational Scalability of the HOPE Block}: By substituting
    quadratic self-attention with the complementary Titans fast-weight mechanism and
    the CMS multi-level consolidation loop, the HOPE block achieves linear
    computational complexity in sequence length. This property is not merely an
    engineering convenience, it directly determines the maximum historical context
    that can be processed during pre-training and, consequently, the richness of the
    representations deposited into the VAE bottleneck.
 
    \item \textbf{Structured Disentanglement as a Representational Prior}: The
    explicit separation of the latent space into trend and seasonal components,
    supervised by the \textit{SeriesDecomp} reconstruction signal, acts as a strong
    inductive prior that guides the encoder toward semantically meaningful partition
    of temporal variation. This prior improves both the interpretability of the
    learned embeddings and their transferability to unseen classification tasks, and
    appears to confer a degree of implicit regularization even in domains where the
    decomposition lacks direct semantic correspondence.
 
    \item \textbf{Efficiency of the Frozen-Encoder Paradigm}: The classification
    results demonstrate that the disentangled VAE bottleneck produces
    sufficiently expressive fixed-length embeddings (formed by concatenating the
    posterior means $[\boldsymbol{\mu}_t \| \boldsymbol{\mu}_s]$ ) for a lightweight
    downstream classifier to achieve competitive accuracy without any encoder update.
    This frozen-encoder protocol is particularly advantageous in low-data regimes,
    where full fine-tuning risks overfitting and erasing the generalist structure
    accumulated during pre-training.
 
    \item \textbf{Domain Alignment Governs Transfer Quality}: The stratification
    of results across dataset types reveals a consistent pattern: domains whose
    temporal structure is compositionally aligned with the trend/seasonal
    factorization (SPECTRO, SIMULATED, HAR, and SENSOR) benefit most from
    ChronoVAE-HOPE's generative prior, while MOTION and DEVICE domains, whose
    high intra-class variability resists clean decomposition, expose the limits of
    the frozen-encoder paradigm. This finding is informative: it localizes the
    performance ceiling not in the HOPE encoder's capacity, but in the
    representational match between the disentanglement prior and the intrinsic
    geometry of the target domain.

    \item \textbf{Performance Ceiling in Multi-Class Scenarios}: Binary classification
    tasks yielded considerably stronger results than multi-class problems, which is
    consistent with the behaviour expected from a fixed-dimensional concatenated
    embedding $[\boldsymbol{\mu}_t \| \boldsymbol{\mu}_s]$ of size $2 \times
    \textit{latent\_dim}$: as the number of classes grows, the probability that a
    linear classifier can partition this compact space into well-separated decision
    regions decreases substantially. Extending the latent dimensionality or
    incorporating a class-conditional prior into the generative process are natural
    directions to address this limitation.
    
    \item \textbf{Information Bottleneck under Extended Sequences}: Performance
    remained stable across short and medium-length time series but deteriorated
    notably beyond 235 time steps. This behaviour is consistent with the fundamental
    tension between the fixed-length nature of the VAE bottleneck and the growing
    information content of longer sequences: as the series lengthens, the compression
    into a single latent vector inevitably discards temporal detail that would be
    necessary for accurate classification. Hierarchical or multi-scale pooling
    strategies constitute the most direct avenue for mitigating this constraint.
    
    \item \textbf{Absence of a Monotonic Relationship with Training Volume}: The
    results across training set size partitions do not conform to the scaling
    behaviour typically associated with data-hungry deep models: accuracy peaks for
    the smallest datasets and deteriorates in intermediate regimes, with only a
    partial recovery at the upper end. Rather than reflecting a direct effect of data
    volume on the frozen encoder representations, this pattern is most plausibly
    explained by the confounding influence of correlated dataset characteristics (such
    as number of classes or sequence length) that co-vary with training set size
    across the benchmark.
    
\end{itemize}

\section{Future Work}
 
Building upon these findings, future lines of research for the evolution of
ChronoVAE-HOPE will focus on the following areas:
 
\begin{enumerate}
    \item \textbf{Disentanglement Regularization via Information-Theoretic
    Constraints}: The current VAE formulation relies on the KL divergence and the
    decomposition reconstruction signal to encourage orthogonality between
    $\mathbf{z}_t$ and $\mathbf{z}_s$. Future work will explore explicit mutual
    information minimization between the two latent subspaces (for instance,
    through total correlation penalties as in $\beta$-TCVAE) to enforce stricter
    statistical independence and further improve the interpretability and
    modularity of the disentangled representations.
 
    \item \textbf{Generative Exploitation for Low-Resource Adaptation}: A key
    advantage of the VAE framework that remains unexploited in the current
    classification protocol is its generative capacity. By sampling from the
    disentangled prior and selectively perturbing the trend or seasonal component
    independently, the decoder can synthesize realistic time series augmentations
    for underrepresented classes. Investigating this generative data augmentation
    loop (particularly for MOTION and DEVICE domains suffering from intra-class
    variability) represents a natural and architecturally coherent next step.
 
    \item \textbf{Adaptive Masked Modeling and Pre-Training Scale}: The MTSM
    objective currently applies a fixed 20\% masking ratio uniformly across
    positions. Curriculum masking strategies (beginning with lower ratios and
    progressively increasing the reconstruction difficulty) alongside expansion of
    the pre-training corpus to larger Monash subsets or additional heterogeneous
    repositories, may accelerate the convergence of the disentangled representations
    and probe whether zero-shot classification emerges as a capability of the
    structured latent space.
 
    \item \textbf{Multivariate Disentanglement}: Extending ChronoVAE-HOPE to
    multivariate inputs requires rethinking the disentanglement strategy: the
    trend and seasonal factorization must account not only for within-channel
    temporal structure but also for cross-channel dependencies that may share
    common trend or seasonal drivers. Architectures based on channel-wise
    posterior factorization with shared prior parameters, or on group-sparse
    latent representations, offer a promising direction for maintaining
    the interpretability of the disentangled bottleneck at higher input
    dimensionality.
\end{enumerate}

\section*{Acknowledgment}

This research has been partially supported by Proyecto PID2023-151336OB-I00 financiado por MICIU/AEI /10.13039/501100011033 y por FEDER, UE.

We do appreciate Prof. Keogh's comments on earlier versions of the paper, which have undoubtedly improved the content. Also, we would like to thank the creators and maintainers of the UCR Time Series Archive \cite{UCRArchive2018} and the Monash Time Series Forecasting Archive for making their comprehensive data repositories publicly available to the research community.

\section*{Appendix A: Complete Results}
\begin{longtable}{l S S}
    \toprule
    \textbf{Dataset} & {\textbf{Accuracy}} & {\textbf{F1-Score}} \\
    \midrule
    \endfirsthead
    
    \toprule
    \textbf{Dataset} & {\textbf{Accuracy}} & {\textbf{F1-Score}} \\
    \midrule
    \endhead
    
    \bottomrule
    \endfoot

    Adiac                          & {14.07} & {7.85}  \\ \hline
    ArrowHead                      & {72.57} & {71.88} \\ \hline
    Beef                           & {23.33} & {22.73} \\ \hline
    BeetleFly                      & {70.00} & {67.03} \\ \hline
    BirdChicken                    & {50.00} & {47.92} \\ \hline
    Car                            & {51.67} & {52.95} \\ \hline
    CBF                            & {91.00} & {90.87} \\ \hline
    ChlorineConcentration          & {54.84} & {40.85} \\ \hline
    CinCECGTorso                   & {30.29} & {29.55} \\ \hline
    Coffee                         & {96.43} & {96.41} \\ \hline
    Computers                      & {64.40} & {64.37} \\ \hline
    CricketX                       & {23.33} & {17.91} \\ \hline
    CricketY                       & {13.33} & {9.63}  \\ \hline
    CricketZ                       & {13.08} & {8.12}  \\ \hline
    DistalPhalanxTW                & {64.03} & {54.18} \\ \hline
    Earthquakes                    & {74.10} & {63.69} \\ \hline
    ECG200                         & {69.00} & {68.90} \\ \hline
    ECG5000                        & {91.47} & {88.86} \\ \hline
    ECGFiveDays                    & {67.71} & {66.82} \\ \hline
    ElectricDevices                & {67.42} & {66.56} \\ \hline
    FaceAll                        & {54.02} & {52.47} \\ \hline
    FaceFour                       & {84.09} & {84.35} \\ \hline
    FacesUCR                       & {47.56} & {40.63} \\ \hline
    Fish                           & {48.00} & {41.08} \\ \hline
    FordA                          & {77.05} & {77.04} \\ \hline
    FordB                          & {61.73} & {61.31} \\ \hline
    GunPoint                       & {79.33} & {79.24} \\ \hline
    Ham                            & {69.52} & {69.09} \\ \hline
    HandOutlines                   & {64.86} & {53.58} \\ \hline
    Haptics                        & {23.70} & {19.62} \\ \hline
    Herring                        & {56.25} & {55.95} \\ \hline
    InlineSkate                    & {16.00} & {11.94} \\ \hline
    InsectWingbeatSound            & {32.83} & {29.28} \\ \hline
    ItalyPowerDemand               & {83.97} & {83.96} \\ \hline
    LargeKitchenAppliances         & {37.87} & {35.87} \\ \hline
    Lightning2                     & {62.30} & {62.23} \\ \hline
    Lightning7                     & {53.42} & {48.23} \\ \hline
    Mallat                         & {33.22} & {22.98} \\ \hline
    Meat                           & {75.00} & {75.35} \\ \hline
    MedicalImages                  & {56.32} & {49.79} \\ \hline
    MiddlePhalanxTW                & {51.95} & {46.30} \\ \hline
    MoteStrain                     & {72.60} & {71.09} \\ \hline
    OliveOil                       & {76.67} & {76.97} \\ \hline
    OSULeaf                        & {27.69} & {23.99} \\ \hline
    PhalangesOutlinesCorrect       & {68.88} & {64.47} \\ \hline
    Plane                          & {92.38} & {92.00} \\ \hline
    ProximalPhalanxTW              & {79.02} & {73.38} \\ \hline
    RefrigerationDevices           & {54.40} & {54.76} \\ \hline
    ScreenType                     & {42.67} & {42.78} \\ \hline
    ShapeletSim                    & {57.22} & {56.63} \\ \hline
    ShapesAll                      & {29.00} & {20.34} \\ \hline
    SmallKitchenAppliances         & {61.07} & {61.18} \\ \hline
    SonyAIBORobotSurface1          & {79.37} & {79.42} \\ \hline
    SonyAIBORobotSurface2          & {81.53} & {81.79} \\ \hline
    Strawberry                     & {85.41} & {85.17} \\ \hline
    SwedishLeaf                    & {75.68} & {75.42} \\ \hline
    SyntheticControl               & {96.67} & {96.65} \\ \hline
    ToeSegmentation1               & {52.63} & {44.17} \\ \hline
    ToeSegmentation2               & {80.00} & {81.04} \\ \hline
    Trace                          & {68.00} & {59.95} \\ \hline
    TwoLeadECG                     & {73.05} & {72.97} \\ \hline
    TwoPatterns                    & {87.28} & {87.25} \\ \hline
    UWaveGestureLibraryX           & {62.70} & {60.24} \\ \hline
    UWaveGestureLibraryY           & {54.55} & {52.94} \\ \hline
    UWaveGestureLibraryZ           & {51.56} & {45.31} \\ \hline
    UWaveGestureLibraryAll         & {62.00} & {59.56} \\ \hline
    Wafer                          & {97.37} & {97.30} \\ \hline
    Wine                           & {74.07} & {72.21} \\ \hline
    WordSynonyms                   & {28.37} & {21.79} \\ \hline
    Worms                          & {40.26} & {38.80} \\ \hline
    WormsTwoClass                  & {61.04} & {60.86} \\ \hline
    Yoga                           & {59.80} & {57.73} \\ \hline
    DistalPhalanxOutlineAgeGroup   & {69.78} & {67.29} \\ \hline
    DistalPhalanxOutlineCorrect    & {72.46} & {70.86} \\ \hline
    MiddlePhalanxOutlineAgeGroup   & {64.94} & {58.73} \\ \hline
    MiddlePhalanxOutlineCorrect    & {76.63} & {76.01} \\ \hline
    NonInvasiveFetalECGThorax1     & {58.93} & {57.46} \\ \hline
    NonInvasiveFetalECGThorax2     & {71.40} & {68.92} \\ \hline
    ProximalPhalanxOutlineAgeGroup & {84.88} & {84.90} \\ \hline
    ProximalPhalanxOutlineCorrect  & {77.66} & {75.66} \\ \hline
    StarLightCurves                & {87.55} & {87.59} \\ \hline

    \caption{ChronoVAE-HOPE results for Bake Off benchmark datasets}
    \label{results}
\end{longtable}

\bibliographystyle{unsrt}
\bibliography{citas}

\end{document}